Exploring the Implementation of AI in Early Onset Interviews to Help Mitigate Bias

Nishka Lal

Omar Benkraouda

New York University

December 18, 2024

# 1. Abstract

This paper investigates the application of artificial intelligence (AI) in early-stage recruitment interviews in order to reduce inherent bias, specifically sentiment bias. Traditional interviewers are often subject to several biases, including interviewer bias, social desirability effects, and even confirmation bias. In turn, this leads to non-inclusive hiring practices, and a less diverse workforce. This study further analyzes various AI interventions that are present in the marketplace today such as multimodal platforms and interactive candidate assessment tools in order to gauge the current market usage of AI in early-stage recruitment. However, this paper aims to use a unique AI system that was developed to transcribe and analyze interview dynamics, which emphasize skill and knowledge over emotional sentiments. Results indicate that AI effectively minimizes sentiment-driven biases by 41.2%, suggesting its revolutionizing power in companies' recruitment processes for improved equity and efficiency.

# 2. Introduction

Though it is a concept that began seemingly a century ago, interviews have become an essential part of the recruitment process. What started as simply as Thomas Edison's filtering mechanism through a lack of worthy college students, has now become the primary part of every companies' HR strategy. With that being said, over the years, there have been constant points upheld about the validity of these interviews, and the implicit biases that are brought with them.

For instance, interview bias is a major source identified by experts in the field. Interviewer bias is basically when personal qualities of the interviewer are key determinants of the outcome of an interview [17]. Furthering this similar idea, Purkiss states that if the interviewer has implicit bias, then even subtle cues may play a role in triggering implicit discrimination responses [20]. For instance, she gives the example of how the U.S. French accents are often associated with sophistication and poise, whereas Asian accents tend to be linked with high economic and educational attainments [20]. Another form of bias that is present is social desirability, meaning that interviewees will often present a more desirable image forward, instead of actually conveying their true sentiments and/or ideas in the interview [17]. In turn, this lack of complete truancy from the candidate leads to the likelihood of underperformance during times of need for the business. Additionally, another form of bias is confirmation bias and the halo effect, which potentially distorts the assessment of candidates and concurrently leads to a workforce that is less varied and inclusive [18].

Altogether, these biases severely undermine the company in the long run. The interviewer bias can lead to a lack of diversity within the company, which then leads to worse problem solving skills, lack of multi-faceted innovation, and unthoughtful strategic thinking [5]. In fact, a study from Harvard Business Review even states that the more similar investment partners are in behaviors and ethnic backgrounds, the consequently lower their investments are as a whole [9]. Furthermore, this bias can lead to the hiring of individuals who are not equitable with the work culture and performance of the company.

With all these biases rising in the new business era, this paper aims to understand, validate, and evaluate the efficiency and applicability of AI to mitigate biases within the recruitment process.

## 3. Literature Review

**3.1- Understanding the Current Market**

Previously, there have been quite a few companies that have attempted to use AI to mitigate biases within the recruitment process. For starters, there was an innovative project started in collaboration with Eni, in which an interface reminiscent of popular social media platforms was provided to the interviewees as an attempt for users to contribute content through various multitudes including text, images, and video [18]. This in turn allows for interviewer bias to be less considered because the median of communication is not just tonal and attitude response, but instead a holistic approach of attempting to understand their overall fit to the company standards. Instead of just one call, it is a series of forums that allow for the candidate to be thoroughly understood, attempting to avoid the first impression in interviewer bias. Another example of AI being used to mitigate bias and time spent in the recruitment process is when Unilever[1] partnered with Pymetrics[2] to create an online platform that uses a candidate's talent, logic, and reasoning, acquired through a neuroscience game event, to screen applicants [13]. Similar to the system Eni has incorporated, this type of system allows skill analysis that is much more holistic than a first impression. It specifically hones in on the candidate's strengths and weaknesses in a more quantitative way, allowing for their characteristics to be matched against those of other people who have proven to be successful in the designated roles (Marr et.al, 2019,132). In fact, this system also can analyze a candidate's facial expressions, gestures, choice of words and body language in front of a mobile phone camera [13]. Overall, this takes away significant interviewer discrimination since their physical features and race are not the prime part of the interview, but instead, it is the talent acquisition. Another mechanism of avoiding bias with AI is by having automated AI models explore and delineate the best fit of a recruiter based on their CV. Basically these recommendation systems use synset grouping and dimension reduction techniques on both the job descriptions and application descriptions to find similarities[2]. In this way, the recruiter's analysis of the candidate's skill is not purely in question, and instead it is more focused on the style of their CV.

---

[1] A consumer goods company that manufactures and sells a variety of products, including food, beverages, beauty and personal care items, and home care products

[2] a game-based assessment tool that helps companies identify the best candidates for a job by evaluating their soft skills

Others have also deemed to focus on not just tangible products and projects, but also smaller ML models that can be used to reduce discrimination bias in the process. For instance, employing Convolutional Neural Networks (CNN)[3] based decision making systems reduces bias and improves accuracy, which can strongly contribute to the proper organization and staff placement [2].

**3.2- Gap/New Solution**

With all that being said, there were not many examples of companies that attempted to mediate the concern of bias in AI by mixing different solutions into one place, enhancing both user experience and comfort. While some of the solutions had just games to consider a different side of the candidate's skills, others analyzed wording on the CV in order to take away initial bias from the interviews.

---

[3] a type of artificial neural network that use three-dimensional data to perform image recognition and processing tasks

However, this paper aims to create and analyze a more specific approach that will allow for companies to have a singular node to refer to when removing bias in initial stages of the interview process. I will then take the new solution into consideration and see if this method performs better or the same as past attempts to reduce biases in early stage recruitment.

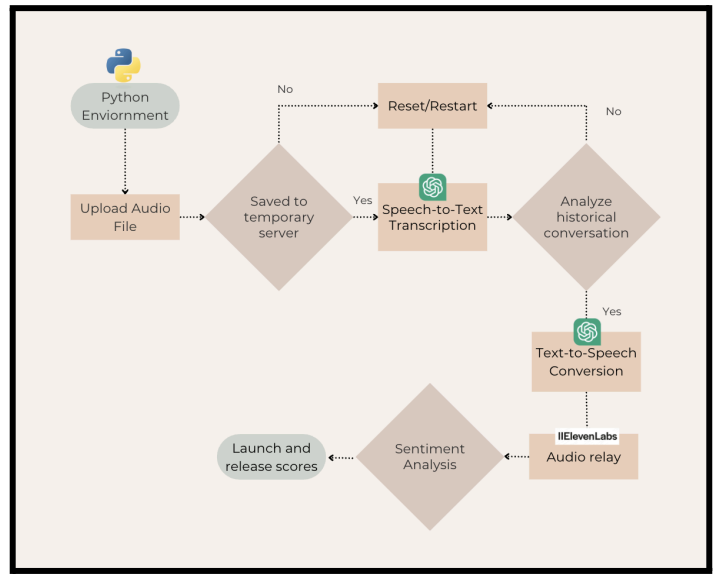

**Figure 1:** Program Workflow Flowchart

This flowchart follows the iterative process of the program that I developed to produce the AI generated ratings

## 4.Methodology

### *4.1 AI Developmental Process*

Before understanding the specific mechanisms that will be used to test if my AI system can accurately reduce levels of bias in early-onset HR interviews, it is important to outline the procedure that was utilized to create the AI interview-bot.

### *4.2 Audio Upload Workflow*

Initially, I used a Python environment in order to begin my code. Within that, I created a main.py file which implements the FastAPI application that I utilized to provide an interactive audio-based conversational system, which I tailored in order to fit the role of a front-end React developer position. The application's primary workflow initialization as the user is asked to upload an audio file through the /talk endpoint. That file is then saved temporarily to a server, and then using OpenAI's Whisper API, the audio is transcribed into text, converting the data to a textual format that can be further processed and analyzed. This aforementioned transcription process is crucial for the workflow of this program because it allows for the audio to be translated into a format that the application can understand, thereby relaying back correct responses to the users comments.

*4.3 Text-to-Speech Response Delivery*

After the transcription, the resulting text is then sent to the OpenAI ChatGPT model [15], which uses the history of the textual conversation and generates a relevant response. This is done through processing the user's input along with accumulated messages on the JSON file, so that not only is the immediate text being reviewed, but so is the background of the said interview. This feature enables the conversation to be more relatable and engaging akin to a face-to-face, instead of disconnected from past talking points. The assistant's response is then converted back to audio form through ElevenLabs text-to-speech APIand is streamed back to the user [8]. This is similar to an actual interview, such that when the user says something, the application responds to it, enabling an interactive experience.

*4.4 Dynamic Sentiment and Reset Features*

Apart from just the conversational functionality, this AI application also incorporates a sentiment analysis feature, which is accessible through the /analyze endpoint. This endpoint allows for the evaluation of the emotional tone in the conversation stored in the JSON file. This sentiment analysis is possible through the utilization of the TextBlob library, which assesses the polarity of the text, categorizing it into positive, negative, or neutral sentiments. This capability allows for users to get an insight into the dynamics present in their conversations, utilizing this to showcase the emotional intelligence of the interviewee, an important skill needed in the workforce.

Finally, the application provides a /clear endpoint, which enables users to reset their chat history as they deem necessary. This feature ensures that users can have a clean slate without any lingering context, so that this application can be used multiple times in a row.

*4.5 Data Collection*

However, the creation of the model was simply the first step. After the model was developed and integrated into the website that I created using Wix, I ran tests on the model in order to gauge a rating scale from the AI itself. Nevertheless, before the data is collected, it is important to define what each of the knowledge levels entail. For the purposes of my study, the ratings are as follows:

<u>Uninformed (1)</u>: Has no knowledge of the topic and does not recognize its relevance, significance, or application.

<u>Basic Awareness (2)</u>: Knows a few terms or steps to begin the process, but lacks understanding. Cannot explain the topic in any meaningful way.

<u>Superficial Understanding (3)</u>: Has a general understanding of the topic but lacks depth regarding the situation or strategy.

<u>Competent (4)</u>: Has a solid understanding of the material and discusses the materials, however not confidently and immediately.

Proficient (5): Possesses in-depth knowledge and expertise, analyzes key ideas well and is able to provide an answer quickly and correctly.

To put each of these into further perspective of the study, a rating of 1 in this case would be if the interviewee demonstrated no understanding of the math problem, not even recognizing the mathematical operations involved in the process. A rating of 2 would be if the interviewee understands the problem and can stack the numbers on top of each other for multiplication, but stops at that point. A rating of 3 would be if the interviewee guesses the right answer by chance or attempts to solve the problem but takes an entirely incorrect path towards doing so. A rating of 4 would be understanding the problem, setting it up correctly and even reaching the correct answer, but having some hesitations and minor doubts in the process. Finally, a rating of 5 would be if the interviewee were able to answer the question not only correctly, but also have limited to no hesitations, and correct steps towards the answer.

Positive sentiment for the purposes of this study relay optimistic toned responses. These responses would be ones that have efforts being made, include positively connotated words, and a pragmatic outlook on failures. Negative sentiment would thereby be the opposite, consisting of a rudimentary tone with shrewd or harsh commentary throughout the iterative process.

Therefore, for each of the tests, I made sure to vary both the knowledge level of the interviewee and their tone. That way, I would be able to see if the sentiment portrayed by the interviewee changed the perception of the interviewee's readiness for the job/task, despite the knowledge levels clearly skewing one way.

I began each interview by first setting up the backend and inputted into my large language model (LLM) the math problem "49*54" so that it would be trained to ask the same initial question each time. I also began each interview inputting the parameters that the AI should be assessing the candidate on, and explaining the number scaling as done above. To display the results, I ended each interview with a question that aimed to retrieve the number from the AI.

For each of the test cases, I varied the sentiment that I portrayed (either positive or negative ) and the knowledge level, and changed my answers to match the designated portfolio for each candidate, as demonstrated by Figure 2.. I made sure to utilize ElevenLabs to use an automated voicebot to speak on my behalf, in order to keep the voice matched and even paced, not adding any extra biases to the data collection. After conducting the entire interview, my AI bot returned the values listed below.

**Figure 2**

*Data Collection for Candidate Interviews*

|  | Knowledge Level (1-5) | Sentiment Analysis | AI Rating (1-5) | Human Rating (1-5) |
|---|---|---|---|---|
| Candidate 1 | 1 | Positive | 1.2 | 3 |
| Candidate 2 | 2 | Positive | 2.5 | 3 |
| Candidate 3 | 3 | Positive | 3.6 | 3.5 |
| Candidate 4 | 4 | Positive | 4 | 4.7 |
| Candidate 5 | 5 | Positive | 4.8 | 5 |
| Candidate 6 | 1 | Negative | 1.5 | 1 |
| Candidate 7 | 2 | Negative | 3 | 2 |
| Candidate 8 | 3 | Negative | 3.4 | 2.3 |
| Candidate 9 | 4 | Negative | 4.5 | 2.7 |
| Candidate 10 | 5 | Negative | 5 | 3 |

Note: This demonstrates the holistic 10 interviews that were conducted and summarizes their ratings and parameters.

However, I not only wanted to see if there was a correlation in sentiment and holistic rating for the AI, but also if there was an added bias that humans displayed. Therefore, I asked two recruiting/hiring managers from Tech Mahindra, one a male and one a female, and asked them to rate the interviews by reading the transcript from the aforementioned interviews. This exact selection was made in order to ensure that there was no added gender bias from the managers, and also to ensure that they had a proficient background, as both of them had 10 or more years of experience in the Human Resources field.

## 5. Discussion

*5.1 Sentiment Bias Calculations/ Data Analysis*

After the data was collected, I deemed to delineate a pattern among my findings. Therefore, I began by analyzing the candidate data for positive sentiments. First, I found the average of the AI ratings was 3.22 and the average of the human ratings which was 3.84. After this, I found the difference between the human and AI averages and found that the difference was 0.62. I repeated the aforementioned process to find the difference in negative sentiment, which was -1.28.

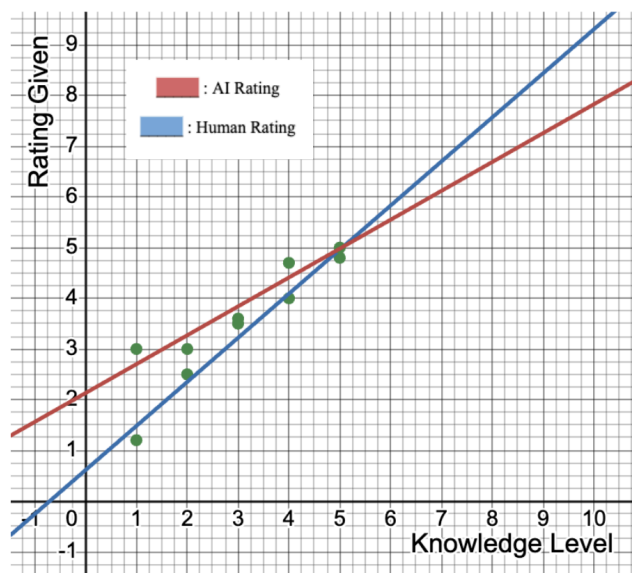

*Figure 3:* This figure demonstrates how the human ratings had a steeper slope, implying that human raters often had a bias towards candidates with positive sentiment driven answers.

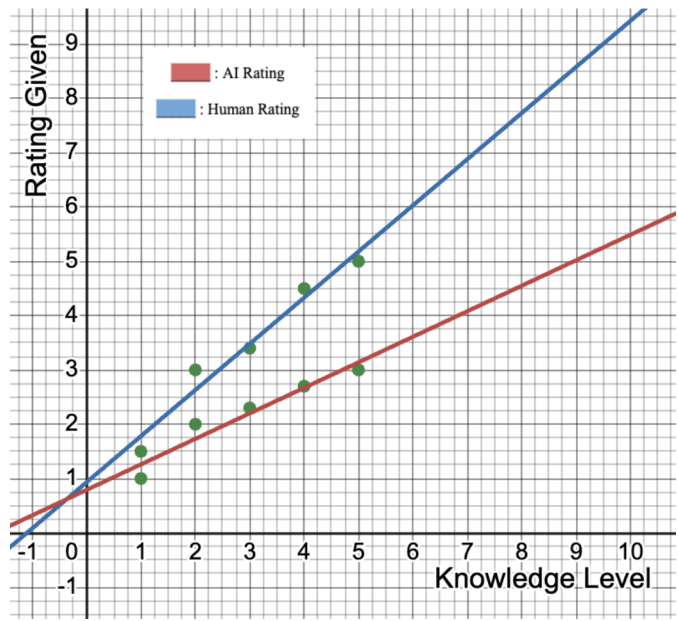

*Figure 4:* This figure demonstrates how the AI ratings had a less steep slope, implying that AI raters often do not react much to negative sentiment in the candidates as compared to human raters.

### 5.2 Findings

From the revelations above, we can deduce that since the difference for positive sentiment was higher, humans tend to favor candidates with positive sentiment more than AI, rating them on an average 0.62 points higher. Furthermore, by looking at the negative bias's high negative difference, we can deduce that humans rate candidates with negative sentiment an average of 1.28 points lower than AI does. Overall, this bias is 2.06 points, highlighting a significant tendency for humans to exhibit a more favorable bias towards candidates with positive sentiment and a consequently harsher assessment of those with negative sentiment compared to my AI bot. Looking at these findings on a more holistic scale, we can deduce that AI bots such as the one that I developed tend to assist remove sentiment driven bias in the early onset recruitment process. Instead of focusing on other subliminal categories of emotion, AI allows for the effective shortlisting of candidates who have capable abilities in their skills and knowledge levels for the job that they must uphold.

*5.3 Limitations*

Nevertheless, while this study has significant additions to the field, there are some minor aspects of this study's process that should be kept in mind. For starters, the methodology section only utilizes 10 test cases since each of them required in depth personal involvement, which may not be a significant amount of data to draw a long-standing conclusion from. Nevertheless, I did aim to make the candidates inclusive of the various types of people in the interviewing situation, but it might be beneficial in a further study to have more test cases run using the AI. Furthermore, using more human interviewers to analyze the interviews might also strengthen the results that were received from the study. Additionally, since this study focuses solely on the importance of knowledge over emotion, these results would not be as effective for jobs that rely heavily on emotional skills. Instead, this study furthers the idea that AI stands as an effective replacement for specifically knowledge/skill based interviews to produce an unbiased recommendation to enhance company productivity.

*5.4 Future Implications*

Furthering upon this research, researchers can expound on using AI to not only mitigate bias in the knowledge/skills section of the interview, but also utilizing its functionalities to minimize gender or even race bias in the emotional aspects as well. Additionally, this study can be implemented in not only mathematical or technical fields, but adapted for career fields that require more niche skill based requirements that are far more physical using features from photo analysis and recognition. At the end of the day, the use of AI in early onset interviews is a growing possibility for the future, and utilizing it to its best potential, whether it be in knowledge/skills sections or a future of more, the possibilities are endless.

## Conclusion

In summary, this paper examines the role of AI in early-stage recruitment interviews, with a specialized focus on reducing sentiment bias. As aforementioned, traditional interview processes are often influenced by interviewer bias, social desirability effects, and confirmation bias, which can lead to detrimental effects such as less inclusive hiring. This study thereby reviews various AI solutions currently available, including multimodal platforms and interactive candidate assessments, to assess the extent to which AI has been integrated in the recruitment domain. Furthermore, this research tested a novel AI system designed to transcribe and evaluate interview dynamics, prioritizing objective assessment of skills and knowledge in comparison to emotional cues. The results demonstrate a 41.2% reduction in sentiment-driven biases, underscoring AI's potential to transform the environment by enhancing efficiency and fairness.

15